\documentclass[runningheads]{llncs}

\usepackage{eccv}

\usepackage{eccvabbrv}

\usepackage{graphicx}
\usepackage{booktabs}
\usepackage{array}
\usepackage{multirow}
\usepackage{amsmath}
\usepackage{amssymb}
\usepackage[table]{xcolor}
\usepackage{subcaption}
\usepackage{pifont}
\usepackage{makecell}
\usepackage{tikz}
\usetikzlibrary{positioning,shapes,arrows.meta}

\usepackage[accsupp]{axessibility}  

\usepackage[breaklinks,colorlinks,citecolor=eccvblue]{hyperref}

\usepackage{orcidlink}

\newcommand{\cmark}{\ding{51}}
\newcommand{\xmark}{\ding{55}}

\begin{document}

\title{FoodMonitor: Benchmarking MLLMs for Explainable Compliance Analysis}

\titlerunning{FoodMonitor: Benchmarking MLLMs for Compliance Analysis}

\author{Ruihao Xu$^{\star}$ \and
Xingming Shui$^{\star}$ \and
Jingxuan Niu$^{\star}$ \and
Yiqin Wang \and
Jilin Yu \and
Haoji Zhang \and
Yansong Tang$^{\dagger}$}

\institute{Tsinghua Shenzhen International Graduate School, Tsinghua University}

\authorrunning{R. Xu, X. Shui, J. Niu et al.}

\maketitle
\let\thefootnote\relax
\footnotetext{\noindent $^{\star}$~Equal contribution. ~$^{\dagger}$~Corresponding author: \url{tang.yansong@sz.tsinghua.edu.cn}}

\begin{abstract}
As AI-powered compliance monitoring becomes increasingly important in public governance and industrial safety, the ability to provide verifiable evidence and traceable accountability signals is essential. However, existing video anomaly detection datasets focus on event-level binary classification, lacking the rule-driven, explainable analysis required for real-world compliance scenarios. We introduce FoodMonitor, a benchmark for explainable compliance analysis in commercial kitchen surveillance. FoodMonitor comprises 477 video clips with 3,307 violation annotations across a dual-channel design covering both person-level and environment-level violations. Each annotation specifies which rule was violated, what non-compliant behavior occurred, and who committed it with frame-level bounding boxes. We establish a unified evaluation protocol with a two-stage matching mechanism that separately assesses spatial localization and semantic understanding, along with a composite metric ($C_{\text{score}}$) that balances environment and person detection performance. Systematic evaluation of several state-of-the-art multimodal large language models reveals that the best-performing model achieves only 0.360 $C_{\text{score}}$, with spatial localization and fine-grained rule understanding emerging as the primary bottlenecks. Our analysis identifies two distinct failure modes: localization-dominated errors and semantics-dominated errors, providing diagnostic insights for future model development.

\end{abstract}

\section{Introduction}
\label{sec:intro}

The deployment of AI systems for compliance monitoring has become increasingly important across domains including public health, workplace safety, and regulatory enforcement. Unlike traditional anomaly detection that produces binary ``normal/abnormal'' classifications, practical compliance monitoring requires systems that can: identify which specific rules are violated, provide evidence explaining why a violation occurred, and attribute violations to responsible individuals when applicable. This combination of rule-driven analysis and explainable outputs is essential for systems that must support human auditors, enable appeals, and maintain accountability.

Video-based compliance monitoring presents unique challenges. In domains such as food safety surveillance, monitors must simultaneously track multiple individuals, recognize subtle violations of specific hygiene requirements, and maintain temporal awareness of behaviors that may constitute violations only in certain contexts. Current video anomaly detection (VAD) benchmarks~\cite{sultani2018ucfcrime,wu2020xdviolence,acsintoae2022ubnormal} focus on event-level anomaly classification without providing the structured, rule-grounded outputs needed for compliance applications. Kitchen video datasets~\cite{damen2018epickitchens,damen2021epickitchens100,tang2019coin} address action recognition rather than regulatory compliance. This gap between academic benchmarks and practical requirements motivates the need for new evaluation frameworks.

The emergence of multimodal large language models (MLLMs) offers potential for more sophisticated compliance analysis. Models such as Gemini~\cite{team2023gemini,team2024gemini15,team2025gemini25}, Qwen3-VL~\cite{bai2025qwen3vl}, GLM-4.6V~\cite{vteam2025glm45v}, MiMo-VL~\cite{xiaomi2025mimovl}, and InternVL3~\cite{zhu2025internvl3} demonstrate remarkable capabilities in visual understanding, reasoning, and structured generation. However, their ability to perform fine-grained, rule-driven compliance analysis with spatial-temporal grounding remains largely unexplored. Existing MLLM video benchmarks~\cite{fu2024videomme,li2024mvbench} evaluate general comprehension through question answering, but do not assess the specific capabilities required for compliance monitoring: mapping observations to predefined rules, generating structured detection outputs, and providing instance-level localization.

\begin{figure}[t]
\centering
\includegraphics[width=\linewidth]{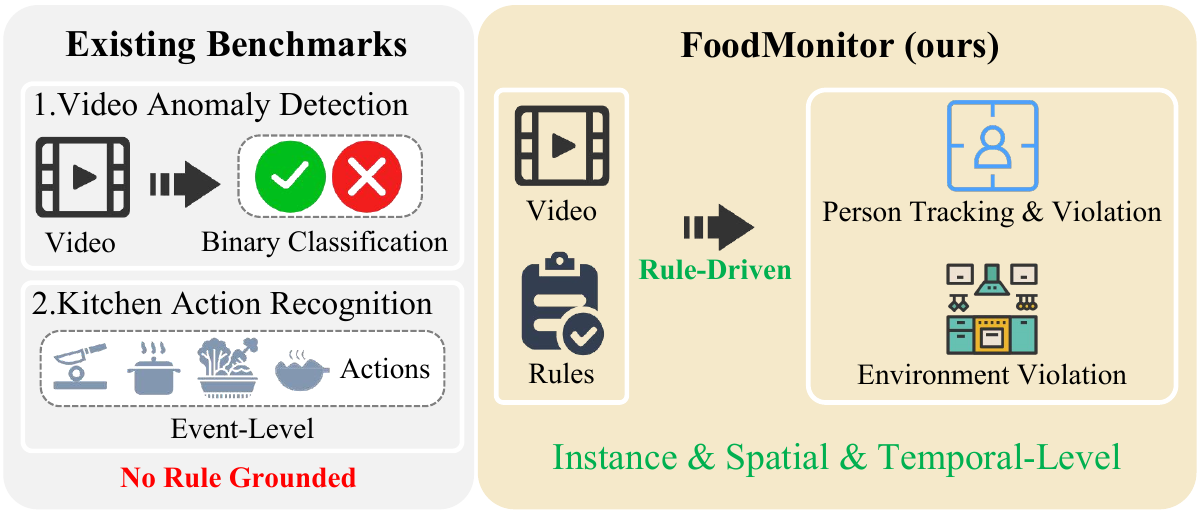}
\caption{Comparison between existing benchmarks and FoodMonitor. Existing video anomaly detection benchmarks produce binary classifications without rule grounding, while kitchen action recognition datasets focus on event-level action labels. FoodMonitor introduces rule-driven compliance analysis that takes video and predefined rules as input, producing instance-level, spatially and temporally grounded outputs for both person violations and environment violations.}
\label{fig:teaser}
\end{figure}

In this paper, we introduce \textbf{FoodMonitor}, a benchmark for explainable compliance analysis in commercial kitchen surveillance. Food safety represents an ideal testbed for compliance monitoring: it involves well-defined regulatory requirements~\cite{fda2017haccp,iso22000}, diverse violation types spanning personal hygiene, food handling, and environmental conditions, and clear accountability needs that require person-level attribution.

FoodMonitor comprises 477 video clips with 3,307 violation annotations organized in a dual-channel structure. The person-level channel captures individual violations such as improper attire, unsafe handling practices, and hygiene breaches, with each annotation specifying which rule was violated, what behavior occurred, and who committed it via frame-level bounding boxes. The environment-level channel addresses facility and equipment violations including sanitation issues, storage problems, and safety hazards. This dual-channel design reflects the practical reality of compliance monitoring, where accountability must be assigned to individuals when applicable while also capturing systemic environmental issues. A five-stage annotation pipeline integrating VLMs, LLMs, and human verification ensures annotation quality and consistency.

We establish a unified evaluation protocol that addresses the unique requirements of compliance analysis. The protocol specifies structured JSON output formats that require models to produce rule-grounded predictions rather than free-form descriptions. A two-stage matching mechanism separately evaluates spatial localization and semantic understanding: the first stage matches predicted person instances to ground-truth annotations using bounding box IoU, while the second stage assesses whether matched predictions correctly identify the violation type. This decomposition enables diagnostic metrics that distinguish between localization failures and semantic failures, providing actionable insights into model weaknesses. The composite $C_{\text{score}}$ metric balances performance across environment and person detection channels.

We evaluate several state-of-the-art MLLMs on FoodMonitor, revealing that even the best performing model achieves only 0.360 $C_{\text{score}}$. Our analysis identifies two distinct failure modes: localization-dominated errors where models fail to correctly associate predictions with ground-truth person instances, and semantic-dominated errors where localization succeeds but violation type identification fails. These findings demonstrate that explainable compliance analysis remains a significant challenge for current MLLMs, highlighting opportunities for future research in spatial perception, fine-grained rule understanding, and reliable structured generation.
Our contributions are summarized as follows:
\begin{itemize}
    \item A comprehensive compliance analysis dataset with 477 videos and 3,307 annotations in a dual-channel structure covering both person-level and environment level violations.
    \item A unified evaluation protocol with structured output formats, two-stage matching, and diagnostic metrics that decompose errors into localization and semantic failures.
    \item Systematic MLLM evaluation revealing current model limitations and distinct failure modes in compliance analysis tasks.
\end{itemize}

\section{Related Work}
\label{sec:related_work}

\subsection{Video Anomaly Detection Datasets}
\label{sec:related_vad_datasets}

Video anomaly detection (VAD) has been driven by increasingly comprehensive benchmark datasets. Early datasets such as UCSD~\cite{hasan2016temporal} and Avenue focused on simple anomalies in constrained environments. UCF-Crime~\cite{sultani2018ucfcrime} introduced real-world surveillance scenarios with 13 anomaly categories, enabling weakly-supervised learning with video-level labels. XD-Violence~\cite{wu2020xdviolence} extended this direction to multimodal violence detection with audio-visual signals. Street Scene~\cite{ramachandra2020streetscene} provided a pixel-level evaluation protocol with diverse anomaly types in urban settings. UBnormal~\cite{acsintoae2022ubnormal} introduced synthetic data for supervised open-set anomaly detection with precise ground-truth annotations.
Kitchen and egocentric video datasets have developed along a separate trajectory. EPIC-KITCHENS \cite{damen2018epickitchens} and its successor EPIC-KITCHENS-100~\cite{damen2021epickitchens100} provide large-scale egocentric recordings of cooking activities with action and object annotations. 50 Salads~\cite{stein201450salads} and COIN~\cite{tang2019coin} focus on procedural activity understanding. Ego4D~\cite{grauman2022ego4d} and Assembly101~\cite{sener2022assembly101} offer diverse egocentric scenarios for various understanding tasks.
However, existing datasets share a fundamental limitation for compliance analysis: they focus on what actions occur (anomaly events, activity labels) rather than whether actions comply with specific regulatory requirements.

\subsection{Video Anomaly Detection Methods}
\label{sec:related_vad_methods}

VAD methods have evolved through several paradigms. Reconstruction-based approaches~\cite{hasan2016temporal,gong2019memorizing,park2020memory} learn to reconstruct normal patterns and detect anomalies via reconstruction error. While effective for simple scenes, these methods struggle with the visual diversity of real-world environments and cannot provide semantic explanations.
Weakly-supervised MIL methods~\cite{sultani2018ucfcrime,feng2021mist,tian2021rtfm,lv2023unbiased} leverage video-level labels to learn discriminative features through multiple instance learning. These approaches achieve strong detection performance but produce only anomaly scores without identifying specific violation types or providing spatial grounding.
Vision-language models have shown strong language-guided visual grounding, e.g., LAVT~\cite{yang2022lavt}.
Recent work has integrated vision-language models into VAD. VadCLIP~\cite{wu2024vadclip} adapts CLIP for weakly-supervised detection using text prompts. Text Prompt with Normality Guidance~\cite{yang2024textprompt} leverages language priors to improve anomaly discrimination. Open-vocabulary approaches~\cite{wu2024openvocab,li2025anomize} extend detection to novel anomaly categories using vision-language alignment. While these methods represent progress toward semantic understanding, they still operate in a detection paradigm that outputs anomaly scores rather than structured compliance assessments with instance-level grounding.

\subsection{MLLMs for Video Understanding}
\label{sec:related_mllm}

Multimodal large language models have rapidly advanced video understanding capabilities. The Gemini family~\cite{team2023gemini,team2024gemini15,team2025gemini25} introduced native multimodal training with million-token context windows, demonstrating strong reasoning across images and videos. Open-source models including Qwen3-VL~\cite{bai2025qwen3vl}, GLM-4.6V~\cite{vteam2025glm45v}, MiMo-VL~\cite{xiaomi2025mimovl}, and InternVL3~\cite{zhu2025internvl3} have progressively closed the gap with proprietary systems.
Benchmarks for MLLM video understanding have also emerged. Video-MME~\cite{fu2024videomme} provides comprehensive evaluation across video types and durations with multiple-choice questions. MVBench~\cite{li2024mvbench} covers 20 temporal understanding tasks. ActivityNet-QA~\cite{yu2019activitynetqa} and EgoSchema~\cite{mangalam2024egoschema} focus on long-form video question answering.
These benchmarks evaluate general video comprehension through question answering, but none addresses the specific requirements of compliance analysis: structured JSON output, rule-category mapping, spatial-temporal instance grounding, and semantic matching against regulatory criteria.


\begin{figure}[t]
\centering
\includegraphics[width=\linewidth]{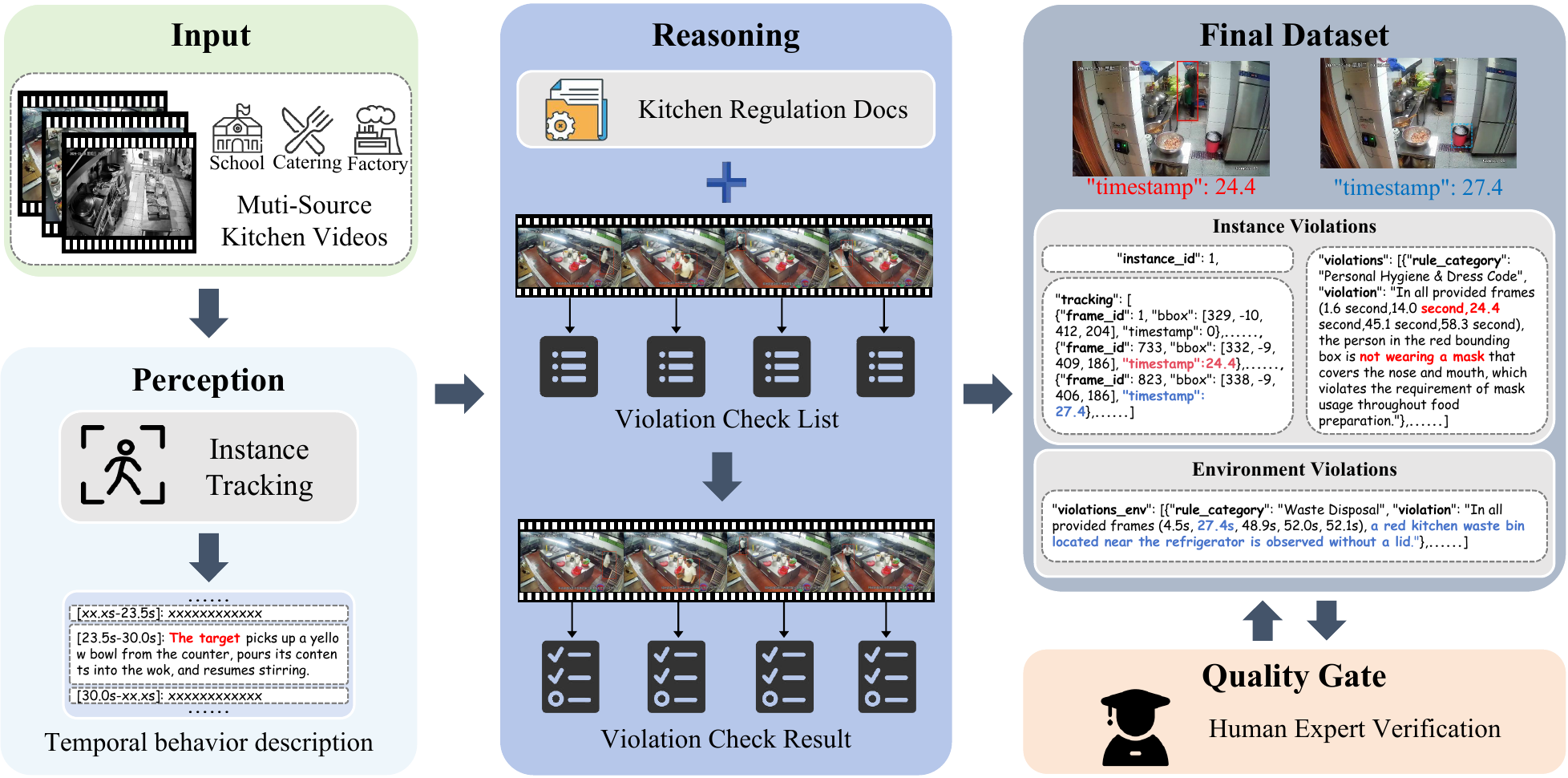}
\caption{Five-stage annotation pipeline for FoodMonitor. The framework integrates automated processing using VLMs and LLMs with human verification to ensure comprehensive coverage of person-level and environment-level violations.}
\label{fig:annotation_pipeline}
\end{figure}

\section{The FoodMonitor Dataset}
\label{sec:dataset}

We present FoodMonitor, a comprehensive benchmark dataset for explainable compliance analysis in commercial kitchen surveillance. Unlike existing video anomaly detection datasets that focus on binary event classification, FoodMonitor provides rule-driven annotations with frame-level person tracking and dual-channel coverage of both person and environment violations.

\subsection{Data Collection}
\label{sec:data_collection}

We collected surveillance videos from diverse commercial kitchen environments including public catering services, school cafeterias, and factory canteens. Each video is processed into standardized 60-second clips at the original frame rate, resulting in 477 evaluation clips. This duration was chosen to balance between capturing complete violation events and maintaining tractable annotation costs.

\subsection{Annotation Pipeline}
\label{sec:annotation_pipeline}

We develop a systematic five-stage annotation pipeline that integrates automated processing with human verification to ensure both scalability and quality. The pipeline addresses the fundamental challenges of temporal reasoning, person-centric tracking, and fine-grained rule verification in complex kitchen surveillance scenarios. \cref{fig:annotation_pipeline} illustrates the overall workflow.

\noindent\textbf{Stage 1: Person Instance Tracking.} We employ Hybrid-SORT~\cite{yang2024hybridsort} to perform multi-object tracking, generating initial pseudo-labels with continuous bounding box trajectories and timestamps for each detected person. Human annotators then verify and refine these trajectories by merging temporally disjoint tracks belonging to the same individual and removing redundant duplicates, yielding accurate person-level spatio-temporal annotations.

\noindent\textbf{Stage 2: Temporal Behavior Captioning.} Based on the refined tracking results, we use a vision-language model to generate temporally-grounded natural language descriptions for each person. The input to the model is a video segment where the target individual is highlighted with a colored bounding box, ensuring the model focuses on the specific person. The generated captions segment actions into time-stamped intervals, providing essential context for identifying violations that depend on temporal continuity, such as cross-contamination events.

\noindent\textbf{Stage 3: Violation Checklist Generation.} We codify food safety regulations into a structured compliance document covering 27 check items across 8 categories (\cref{tab:rule_categories}). A large language model analyzes the temporal behavior captions against these codified regulations to identify potential violations, generating structured checklist items that specify: (1) what to verify, (2) when to check, and (3) which regulation category applies.

\noindent\textbf{Stage 4: Visual Verification and Environment Assessment.} This stage performs two parallel tasks. For person-level violations, the system extracts video frames sampled within the temporal ranges specified in Stage 3, with the target person highlighted by bounding boxes. A VLM examines these frames to provide structured judgments with detailed explanations. For environment-level violations, the system samples random frames throughout the video and performs holistic scene analysis to detect issues such as floor cleanliness, surface sanitation, and waste disposal compliance.

\noindent\textbf{Stage 5: Human Quality Assurance.} Expert annotators review all automated annotations, correcting errors and ensuring consistency with the codified regulations. This final verification step is crucial for maintaining annotation quality and reliability.

\subsection{Annotation Structure}
\label{sec:annotation_structure}

The FoodMonitor dataset employs a hierarchical annotation schema designed to support multi-granularity compliance analysis. \cref{tab:annotation_structure} details the complete structure.

\begin{table}[t]
\centering
\caption{Hierarchical annotation structure of FoodMonitor. The schema supports video-level metadata, frame-level person tracking, and structured violation records for both person and environment channels.}
\label{tab:annotation_structure}
\small
\begin{tabular}{@{}l|l|l|p{5.8cm}@{}}
\toprule
\textbf{Level} & \textbf{Field} & \textbf{Type} & \textbf{Description} \\
\midrule
\multirow{5}{*}{Video}
& \texttt{id} & str & Unique video identifier \\
& \texttt{video\_path} & str & Relative path to video file \\
& \texttt{duration} & int & Video duration (60 seconds) \\
& \texttt{instance\_annotation} & list & Person instance annotations \\
& \texttt{violations\_env} & list & Environment-level violations \\
\midrule
\multirow{3}{*}{\shortstack[l]{Person\\Instance}}
& \texttt{instance\_id} & int & Unique person identifier \\
& \texttt{tracking} & list & Frame-by-frame $(t, \text{bbox})$ pairs \\
& \texttt{violations} & list & Person-level violation records \\
\midrule
\multirow{2}{*}{Violation}
& \texttt{rule\_category} & str & Food safety regulation category \\
& \texttt{violation} & str & Natural language violation description \\
\bottomrule
\end{tabular}
\end{table}

A key design choice is the \textbf{dual-channel} violation structure. Person-level violations (\texttt{violations}) capture individual non-compliant behaviors such as improper mask wearing or food handling errors, with each violation linked to specific person instances through the tracking data. Environment-level violations (\texttt{violations\_env}) capture scene-wide compliance issues independent of specific individuals, such as uncovered waste bins or dirty work surfaces. This separation enables targeted analysis of different violation sources and supports practical deployment scenarios where person identification may be required for accountability.

\begin{table}[t]
\centering
\caption{Rule categories and check item distribution in FoodMonitor. Person-related categories cover individual behavior compliance while environment categories address scene-wide conditions.}
\label{tab:rule_categories}
\small
\newcolumntype{L}[1]{>{\raggedright\arraybackslash}p{#1}}
\begin{tabular}{@{}L{2.5cm}|L{8cm}|r@{}}
\hline
\textbf{Rule Category} & \textbf{Check Items} & \textbf{Count} \\
\hline
\rowcolor{gray!20}
\multicolumn{3}{@{}l}{\textit{Person-Related Violations}} \\
\hline
\makecell[lt]{Personal Hygiene \\ \& Dress Code} & uniform; hat/hair net; hair coverage; mask wearing; mask position & 2,381 \\
\hline
\makecell[lt]{Food Handling \\ \& Processing} & raw/cooked separation; cutting board/knife hygiene; container use; floor placement; sanitary location; garbage handling; cloth cross-contamination & 145 \\
\hline
Hand Hygiene & hand washing after raw food/garbage/restroom; glove use for ready-to-eat food; glove change; hand washing after touching face & 3 \\
\hline
Personal Habits & smoking; chasing/roughhousing; eating/drinking in processing area & 4 \\
\hline
\rowcolor{gray!20}
\multicolumn{3}{@{}l}{\textit{Environment-Related Violations}} \\
\hline
\makecell[lt]{Work Surfaces \\ \& Shelves} & surface cleanliness; shelf organization; personal items placement & 412 \\
\hline
Waste Disposal & waste bin lids; overflow management & 278 \\
\hline
\makecell[lt]{Floor \& Wall \\ Cleanliness} & standing water/grease/residue; stains/mold/cobwebs & 44 \\
\hline
Cleaning Tools & tool placement proximity to food areas & 40 \\
\hline
\multicolumn{2}{@{}l|}{\textbf{Total (27 items)}} & \textbf{3,307} \\
\hline
\end{tabular}
\end{table}

\subsection{Dataset Statistics}
\label{sec:dataset_statistics}

\noindent\textbf{Overview.} The FoodMonitor evaluation set comprises 477 MP4 video clips, each 60 seconds in duration.

\noindent\textbf{Instance Density.} The dataset contains 1,031 person instances with an average of 2.16 instances per clip. Importantly, 78.4\% of clips contain multiple persons (374 out of 477), reflecting the complexity of real commercial kitchens where multiple workers operate simultaneously. This multi-person characteristic poses significant challenges for models that must correctly associate violations with specific individuals.

\noindent\textbf{Violation Distribution.} The 3,307 violations exhibit a highly imbalanced distribution across categories. Personal Hygiene \& Dress Code dominates with 2,381 violations (72.0\%), primarily due to mask-related infractions. In contrast, Hand Hygiene and Personal Habits categories contain only 7 violations combined, representing rare but critical safety concerns. This long-tail distribution mirrors real-world compliance scenarios and challenges models to maintain sensitivity across both common and rare violation types.

\noindent\textbf{Comparison with Existing Datasets.} \cref{tab:dataset_comparison} compares FoodMonitor with related datasets. Unlike video anomaly detection benchmarks (UCF-Crime~\cite{sultani2018ucfcrime}, XD-Violence~\cite{wu2020xdviolence}) that provide only video-level or segment-level binary labels, FoodMonitor offers rule-grounded violation descriptions with instance-level spatial tracking. Unlike kitchen/egocentric datasets (EPIC-KITCHENS~\cite{damen2018epickitchens}, 50 Salads~\cite{stein201450salads}) that focus on action recognition, FoodMonitor specifically addresses compliance verification against codified regulations. Commercial food safety monitoring systems typically operate on proprietary data; FoodMonitor fills this gap by providing a publicly available benchmark with comprehensive annotations.

\begin{table}[t]
\centering
\caption{Comparison of FoodMonitor with existing datasets. FoodMonitor uniquely combines rule-driven compliance analysis with instance-level tracking in a publicly available benchmark.}
\label{tab:dataset_comparison}
\small
\begin{tabular}{@{}l|ccccc@{}}
\toprule
\textbf{Capability} & \rotatebox{50}{\textbf{UCF-Crime}} & \rotatebox{50}{\textbf{XD-Violence}} & \rotatebox{50}{\textbf{EPIC-KITCHENS}} & \rotatebox{50}{\textbf{50 Salads}} & \rotatebox{50}{\textbf{FoodMonitor}} \\
\midrule
Rule-based check items & \xmark & \xmark & \xmark & \xmark & \cmark \\
Instance-level tracking & \xmark & \xmark & \xmark & \xmark & \cmark \\
Violation time localization & \xmark & \cmark & \xmark & \xmark & \cmark \\
Multi-category coverage & \xmark & \cmark & \cmark & \xmark & \cmark \\
Explainable evidence & \xmark & \xmark & \xmark & \xmark & \cmark \\
Publicly available & \cmark & \cmark & \cmark & \cmark & \cmark \\
\bottomrule
\end{tabular}
\end{table}

\section{Benchmark Protocol}
\label{sec:benchmark}

We establish a unified evaluation protocol for explainable compliance analysis that addresses the unique challenges of this task: structured output generation, spatial-temporal grounding, and semantic rule matching. The protocol consists of four components: task definition, input/output specification, matching mechanism, and metric design.

\subsection{Task Definition}
\label{sec:task_definition}

We formulate the task as \textbf{Detection-based Explainable Compliance Analysis}: given a kitchen surveillance video and a compliance rule document, the model must identify all violations present in the video, providing for each violation: (1) the applicable rule category, (2) a natural language description of the observed non-compliant behavior or condition, and (3) for person-level violations, spatial-temporal anchors that localize the violating individual.

This formulation differs from traditional video anomaly detection in several key aspects. First, it requires rule-grounded reasoning rather than open-ended anomaly identification---models must map observations to specific regulatory requirements. Second, it demands structured output including explicit rule category assignment, rather than simple binary or confidence scores. Third, for person violations, it requires instance-level grounding through spatial bounding boxes and temporal timestamps, enabling accountability tracing.

\subsection{Input/Output Protocol}
\label{sec:io_protocol}

\noindent\textbf{Input.} Each model receives: (1) uniformly sampled video frames at 1.0 FPS, and (2) the complete compliance rule document containing all 27 check items organized by category.

\noindent\textbf{Output.} Models produce structured predictions in two channels:
For \textit{environment-level violations}, each detection specifies a rule category and a natural language description of the observed issue. For \textit{person-level violations}, each detection additionally includes a sequence of spatial-temporal anchors $\{(t_k, [x_1, y_1, x_2, y_2])\}_{k=1}^{K}$ ($K \geq 10$) that localize the violating individual via bounding boxes across frames.

The anchor requirement ensures sufficient evidence for instance-level matching. Malformed outputs are handled through automatic retry with error feedback.

\subsection{Matching Protocol}
\label{sec:matching_protocol}

Evaluating compliance detection requires matching model predictions against ground-truth annotations. We design separate matching procedures for environment and person channels to address their distinct characteristics.

\noindent\textbf{Environment Violation Matching.} For environment detections, matching proceeds in two steps. \textit{Step 1: Category Filtering.} Given a prediction $d^e = (c^e, v^e)$ with category $c^e$ and violation description $v^e$, we first filter the ground-truth set $\mathcal{G}_{\text{env}}$ to obtain candidates with matching categories: $\mathcal{G}^{c}_{\text{env}} = \{g \in \mathcal{G}_{\text{env}} \mid g.\text{category} = c^e\}$. \textit{Step 2: Semantic Matching.} We invoke a text comparison model to determine whether $v^e$ semantically matches any candidate in $\mathcal{G}^{c}_{\text{env}}$. A match occurs if the descriptions refer to the same environmental issue, allowing for paraphrase variation. Predictions that match a ground-truth entry are true positives (TP); unmatched predictions are false positives (FP); unmatched ground-truth entries are false negatives (FN). We enforce one-to-one matching constraints: each ground-truth entry can be matched by at most one prediction, preventing inflated scores from redundant detections.

\noindent\textbf{Person Violation Matching.} Person violations require an additional spatial-temporal localization step before semantic matching. The two-stage procedure is:

\noindent \textit{Stage 1: Instance Localization.} For each person prediction $d^p$ with anchor set $\mathcal{A} = \{(t_k, b_k)\}_{k=1}^{K}$, we compute its alignment with each ground-truth person instance $j$ having tracking sequence $\mathcal{T}_j = \{(t_l^{\text{gt}}, b_l^{\text{gt}})\}$. For each anchor $(t_k, b_k)$, we find the temporally closest ground-truth tracking point within tolerance $\tau_t = 0.5$s and compute the Intersection-over-Union (IoU):
\begin{equation}
\text{IoU}(b_k, b_l^{\text{gt}}) = \frac{|b_k \cap b_l^{\text{gt}}|}{|b_k \cup b_l^{\text{gt}}|}
\end{equation}
If no tracking point falls within the temporal tolerance, IoU is set to 0 for that anchor. We then compute aggregate metrics for each candidate instance:
\begin{equation}
\bar{s}_j = \frac{1}{K}\sum_{k=1}^{K} \text{IoU}_k^{(j)}, \quad \rho_j = \frac{|\{k : \exists l, |t_l^{\text{gt}} - t_k| \leq \tau_t\}|}{K}
\end{equation}
where $\bar{s}_j$ is the mean IoU score and $\rho_j$ is the temporal coverage ratio. The prediction is assigned to instance $j^* = \arg\max_j \bar{s}_j$. Localization succeeds if $\bar{s}_{j^*} \geq \theta_s$ (IoU threshold, default 0.3) and $\rho_{j^*} \geq \theta_\rho$ (coverage threshold, default 0.6).

\noindent \textit{Stage 2: Semantic Matching.} This stage is only executed for predictions that successfully pass Stage 1. For these localized predictions, we perform category filtering and semantic matching against the global person violation set $\mathcal{G}_{\text{per}}$, using the same procedure as environment matching with one-to-one constraints. Predictions that fail localization are immediately classified as FP without entering semantic matching.

The final classification of person predictions is:
\begin{equation}
d^p \mapsto \begin{cases}
\text{TP} & \text{if localization succeeds AND semantic match found} \\
\text{FP} & \text{otherwise}
\end{cases}
\end{equation}
False negatives (FN) are counted as ground-truth violations that remain unmatched after processing all predictions.

This two-stage design enables diagnostic analysis: FP predictions can be categorized as localization failures (failed Stage 1) or semantic failures (passed Stage 1 but failed Stage 2).

\subsection{Metrics}
\label{sec:metrics}

\noindent\textbf{Primary Metrics.} For each channel $x \in \{\text{env}, \text{per}\}$, we compute standard detection metrics:
\begin{equation}
P_x = \frac{\text{TP}_x}{\text{TP}_x + \text{FP}_x}, \quad
R_x = \frac{\text{TP}_x}{\text{TP}_x + \text{FN}_x}, \quad
F_{1,x} = \frac{2 P_x R_x}{P_x + R_x}
\end{equation}

The overall benchmark score combines both channels with equal weight:
\begin{equation}
C_{\text{score}} = \frac{1}{2}(F_{1,\text{env}} + F_{1,\text{per}})
\end{equation}

\noindent\textbf{Diagnostic Metrics for Person Detection.} To enable failure mode analysis, we report additional process metrics specific to the person channel, which involves the two-stage matching procedure:
\begin{itemize}
    \item \textbf{Instance Match Rate (IMR)}: $N_{\text{loc}} / N_{\text{pred}}$, the fraction of predictions that successfully localize to a ground-truth instance.
    \item \textbf{Semantic Hit Rate (SHR)}: $N_{\text{hit}} / N_{\text{loc}}$, the fraction of localized predictions that achieve semantic match.
    \item \textbf{FP Localization Ratio} ($r_{\text{loc}}$): $N_{\text{FP,loc}} / N_{\text{FP}}$, the proportion of false positives due to localization failure.
    \item \textbf{FP Semantic Ratio} ($r_{\text{sem}}$): $N_{\text{FP,sem}} / N_{\text{FP}}$, the proportion of false positives due to semantic mismatch.
\end{itemize}

These metrics decompose model errors into spatial perception failures ($r_{\text{loc}}$) and rule understanding failures ($r_{\text{sem}}$), providing actionable insights for model improvement.

\begin{table}[t]
\centering
\caption{Main benchmark results on FoodMonitor. All models are evaluated in thinking mode. Act.: activated parameters; P: Precision; R: Recall. \textbf{Bold} indicates best, \underline{underline} indicates second-best. Doubao-Seed-2.0-Pro leads with $C_{\text{score}} = 0.360$, but absolute performance remains low. All models show precision-dominant performance with relatively low recall, indicating conservative prediction behavior.}
\label{tab:main_leaderboard}
\small
\begin{tabular}{@{}ll|c|ccc|ccc@{}}
\toprule
& & & \multicolumn{3}{c|}{\textbf{Environment}} & \multicolumn{3}{c}{\textbf{Person}} \\
\textbf{Model} & \textbf{Act.} & $\mathbf{C_{\text{score}}}$ & P & R & F1 & P & R & F1 \\
\midrule
\rowcolor{gray!20} \multicolumn{9}{@{}l}{\textit{Closed-source}} \\
Doubao-Seed-2.0-Pro & -- & \textbf{0.360} & \underline{0.572} & \underline{0.384} & \textbf{0.459} & \textbf{0.335} & \textbf{0.214} & \textbf{0.261} \\
Doubao-Seed-2.0-Lite & -- & 0.298 & 0.564 & 0.336 & \underline{0.421} & 0.273 & 0.129 & 0.175 \\
Gemini-3-Pro & -- & 0.250 & 0.471 & 0.217 & 0.297 & 0.250 & 0.171 & 0.203 \\
Gemini-3-Flash & -- & \underline{0.316} & 0.510 & 0.297 & 0.376 & \underline{0.332} & \underline{0.208} & \underline{0.256} \\
\rowcolor{gray!20} \multicolumn{9}{@{}l}{\textit{Open-source}} \\
GLM-4.6V & 106B & 0.279 & 0.323 & \textbf{0.466} & 0.382 & 0.236 & 0.141 & 0.177 \\
GLM-4.6V-Flash & 9B & 0.135 & 0.265 & 0.222 & 0.242 & 0.052 & 0.019 & 0.028 \\
Qwen3.5-397B-A17B & 17B & 0.231 & 0.443 & 0.252 & 0.321 & 0.198 & 0.108 & 0.140 \\
Qwen3-VL-235B-A22B-Thinking & 22B & 0.193 & 0.463 & 0.225 & 0.303 & 0.108 & 0.069 & 0.084 \\
Qwen3-VL-8B-Thinking & 8B & 0.130 & 0.554 & 0.125 & 0.204 & 0.072 & 0.045 & 0.056 \\
Kimi-K2.5 & 32B & 0.208 & 0.366 & 0.194 & 0.254 & 0.195 & 0.141 & 0.163 \\
MiMo-VL-7B-RL & 7B & 0.132 & \textbf{0.633} & 0.105 & 0.180 & 0.135 & 0.060 & 0.084 \\
\bottomrule
\end{tabular}
\end{table}

\section{Experiments}
\label{sec:experiments}

We conduct systematic evaluation of state-of-the-art multimodal large language models (MLLMs) on the FoodMonitor benchmark to assess their capabilities in explainable compliance analysis. Our evaluation covers 11 representative models spanning both closed-source and open-source families, tested under unified protocols with consistent input formats and evaluation metrics. 
Through detailed error decomposition, we identify two distinct failure modes, providing actionable insights for future model development in video-based compliance monitoring.

\subsection{Experimental Setup}
\label{sec:exp_setup}

\noindent\textbf{Evaluated Models.} We evaluate a diverse set of MLLMs with video understanding capabilities, spanning both closed-source and open-source families. Closed-source models include Doubao-Seed-2.0 series~\cite{bytedance2026doubao}, Gemini-3-Flash and Gemini-3-Pro~\cite{team2024gemini15}. Open-source models include GLM-4.6V series~\cite{zhipuai2025glm46v}, Kimi-K2.5~\cite{moonshotai2025kimi}, Qwen3-VL and Qwen3.5 series~\cite{bai2025qwen25vl}, and MiMo-VL-7B~\cite{xiaomi2025mimovl}.

\noindent\textbf{Evaluation Configuration.} Each model receives the same input: 60-second videos sampled at 1.0 FPS (60 frames), encoded as base64, along with the complete 27-item rule document. We use consistent prompts specifying the required JSON output format, coordinate conventions, and minimum anchor requirements. The matching thresholds are set to $\theta_s = 0.3$ (IoU) and $\theta_\rho = 0.6$ (coverage), with temporal tolerance $\tau_t = 0.5$s.

\subsection{Main Results}
\label{sec:main_results}

\cref{tab:main_leaderboard} presents the main benchmark results. The best-performing model, Doubao-Seed-2.0-Pro, achieves only $C_{\text{score}} = 0.360$, demonstrating that explainable compliance analysis remains a significant challenge for current MLLMs. Additionally, we highlight three key findings.

\noindent\textbf{Environment vs.\ Person Detection.} All models perform substantially better on environment violations than person violations. This gap reflects the additional difficulty of person-level detection, which requires accurate spatial-temporal localization in addition to semantic understanding.

\noindent\textbf{Precision vs.\ Recall Trade-off.} Most models exhibit higher precision than recall on both channels, indicating conservative prediction behavior where models tend to predict violations only when highly confident, resulting in fewer but more accurate predictions at the cost of missing some true violations.

\noindent\textbf{Closed-source vs.\ Open-source Gap.} Closed-source models generally outperform open-source alternatives, with the top three $C_{\text{score}}$ values achieved by Doubao-Seed-2.0-Pro (0.360), Gemini-3-Flash (0.316), and Doubao-Seed-2.0-Lite (0.298). However, the best open-source model (GLM-4.6V, 0.279) achieves competitive performance, demonstrating the rapid progress of open-source MLLMs.

\begin{table}[t]
\centering
\caption{Person detection error decomposition. All models are evaluated in thinking mode. IMR and SHR measure success rates at each matching stage. $r_{\text{loc}}$ and $r_{\text{sem}}$ decompose false positives by failure type.}
\label{tab:error_decomposition}
\small
\begin{tabular}{@{}ll|cc|cc@{}}
\toprule
\textbf{Model} & \textbf{Act.} & \textbf{IMR}$\uparrow$ & \textbf{SHR}$\uparrow$ & $\mathbf{r_{\text{loc}}}$$\downarrow$ & $\mathbf{r_{\text{sem}}}$$\uparrow$ \\
\midrule
\rowcolor{gray!20} \multicolumn{6}{@{}l}{\textit{Closed-source}} \\
Doubao-Seed-2.0-Pro & -- & \textbf{0.729} & 0.459 & \textbf{0.407} & \textbf{0.593} \\
Doubao-Seed-2.0-Lite & -- & \underline{0.695} & 0.393 & \underline{0.420} & \underline{0.580} \\
Gemini-3-Flash & -- & 0.580 & 0.573 & 0.629 & 0.371 \\
Gemini-3-Pro & -- & 0.416 & 0.600 & 0.778 & 0.222 \\
\rowcolor{gray!20} \multicolumn{6}{@{}l}{\textit{Open-source}} \\
GLM-4.6V & 106B & 0.385 & 0.614 & 0.805 & 0.195 \\
GLM-4.6V-Flash & 9B & 0.158 & 0.329 & 0.888 & 0.112 \\
Qwen3.5-397B-A17B & 17B & 0.370 & 0.535 & 0.785 & 0.215 \\
Qwen3-VL-235B-A22B-Thinking & 22B & 0.211 & 0.512 & 0.884 & 0.116 \\
Qwen3-VL-8B-Thinking & 8B & 0.109 & \textbf{0.665} & 0.961 & 0.039 \\
Kimi-K2.5 & 32B & 0.378 & 0.515 & 0.773 & 0.227 \\
MiMo-VL-7B-RL & 7B & 0.210 & \underline{0.646} & 0.914 & 0.086 \\
\bottomrule
\end{tabular}
\end{table}

\subsection{Error Analysis}
\label{sec:error_analysis}

We leverage the diagnostic metrics from \cref{sec:metrics} to decompose person detection errors into localization and semantic failures. \cref{tab:error_decomposition} presents the results.

\noindent\textbf{Two Distinct Failure Modes.} The error decomposition reveals two contrasting patterns:
\textbf{\textit{Localization-dominated failures}}: Model predictions frequently fail to align with any ground-truth person instance, exhibiting low Instance Match Rate (IMR $<$ 0.42) and high localization FP ratio ($r_{\text{loc}} > 0.77$), indicating fundamental difficulties in spatial perception and person tracking. Representative models include Gemini-3-Pro, GLM-4.6V, and Qwen3-VL-8B-Thinking.
\textbf{\textit{Semantic-dominated failures}}: Models excel at spatial perception and successfully localize most predictions to ground-truth instances, but frequently misidentify the type of violation---predicting violations that don't semantically match the ground-truth labels, exhibiting high IMR but relatively lower semantic hit rate and higher semantic FP ratio. Representative models include Doubao-Seed-2.0-Pro and Doubao-Seed-2.0-Lite.

\begin{figure}[t]
\centering
\includegraphics[width=0.9\linewidth]{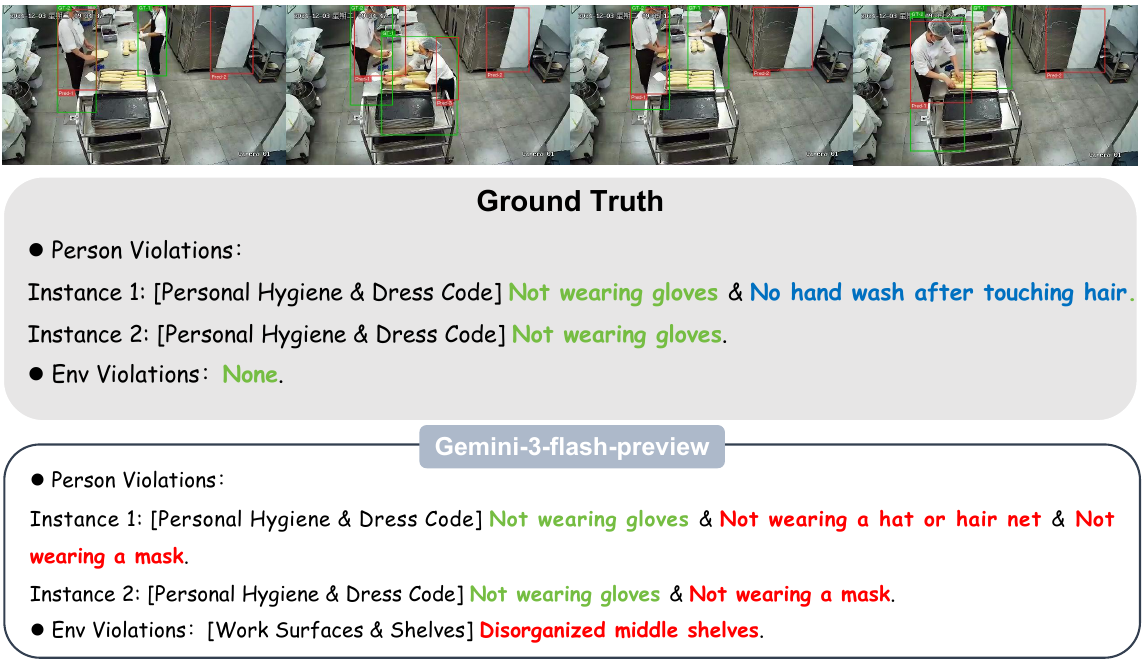}
\caption{Case study comparing ground-truth annotations with model predictions. Green boxes indicate true positives, red boxes denote false positives, and blue boxes highlight false negatives.}
\label{fig:case_study}
\end{figure}

\noindent\textbf{Spatial Localization as Primary Bottleneck.} According to the error decomposition metrics, localization failures account for the majority of false positives across most models. This finding suggests that improving spatial perception and person tracking should be a priority for future model development in compliance analysis tasks. \cref{fig:case_study} illustrates representative examples of these failure modes.

\section{Conclusion}
\label{sec:conclusion}

We presented FoodMonitor, a benchmark for explainable compliance analysis in commercial kitchen surveillance, comprising 477 videos with 3,307 dual-channel annotations and a two-stage matching protocol that decomposes errors into localization and semantic failures.
Evaluation of state-of-the-art MLLMs reveals that the best model achieves only 0.360 $C_{\text{score}}$, with spatial localization as the primary bottleneck across most models. We hope FoodMonitor will advance MLLMs toward practical, accountable compliance monitoring.

\bibliographystyle{splncs04}
\bibliography{main}

\begin{thebibliography}{10}
\providecommand{\url}[1]{\texttt{#1}}
\providecommand{\urlprefix}{URL }
\providecommand{\doi}[1]{https://doi.org/#1}

\bibitem{acsintoae2022ubnormal}
Acsintoae, A., Florescu, A., Georgescu, M.I., Mare, T., Sumedrea, P., Ionescu, R.T., Khan, F.S., Shah, M.: Ubnormal: New benchmark for supervised open-set video anomaly detection. In: CVPR. pp. 20111--20121 (2022)

\bibitem{bai2025qwen3vl}
Bai, S., Cai, Y., Chen, R., Chen, K., et~al.: Qwen3-vl technical report. arXiv preprint arXiv:2511.21631  (2025)

\bibitem{bai2025qwen25vl}
Bai, S., Chen, K., Liu, X., Wang, J., Ge, W., Song, S., Dang, K., Wang, P., Wang, S., Tang, J., et~al.: Qwen2.5-vl technical report. arXiv preprint arXiv:2502.13923  (2025)

\bibitem{bytedance2026doubao}
{ByteDance Seed}: Seed2.0 model card: Towards intelligence frontier for real-world complexity (2026), \url{https://seed.bytedance.com/en/blog/seed2-0-release}, official release page

\bibitem{damen2018epickitchens}
Damen, D., Doughty, H., Farinella, G.M., Fidler, S., Furnari, A., Kazakos, E., Moltisanti, D., Munro, J., Perrett, T., Price, W., Wray, M.: Scaling egocentric vision: The epic-kitchens dataset. In: ECCV. pp. 753--771 (2018)

\bibitem{damen2021epickitchens100}
Damen, D., Doughty, H., Farinella, G.M., Furnari, A., Kazakos, E., Ma, J., Moltisanti, D., Munro, J., Perrett, T., Price, W., Wray, M.: Rescaling egocentric vision: Collection, pipeline and challenges for epic-kitchens-100. IJCV  \textbf{130}(1),  33--55 (2022). \doi{10.1007/s11263-021-01531-2}

\bibitem{feng2021mist}
Feng, J.C., Hong, F.T., Zheng, W.S.: Mist: Multiple instance self-training framework for video anomaly detection. In: CVPR. pp. 14009--14018 (2021)

\bibitem{fu2024videomme}
Fu, C., Dai, Y., Luo, Y., Li, L., Ren, S., Zhang, R., Wang, Z., Zhou, C., Shen, Y., Zhang, M., et~al.: Video-mme: The first-ever comprehensive evaluation benchmark of multi-modal llms in video analysis. In: CVPR. pp. 24108--24118 (2025). \doi{10.1109/CVPR52734.2025.02245}

\bibitem{team2025gemini25}
{Gemini Team}: Gemini 2.5: Pushing the frontier with advanced reasoning, multimodality, long context, and next generation agentic capabilities. arXiv preprint arXiv:2507.06261  (2025), \url{https://arxiv.org/abs/2507.06261}

\bibitem{team2023gemini}
{Gemini Team, Google}: Gemini: A family of highly capable multimodal models. arXiv preprint arXiv:2312.11805  (2023)

\bibitem{team2024gemini15}
{Gemini Team, Google}: Gemini 1.5: Unlocking multimodal understanding across millions of tokens of context. arXiv preprint arXiv:2403.05530  (2024)

\bibitem{vteam2025glm45v}
{GLM-V Team}, Hong, W., Yu, W., Gu, X., Wang, G., et~al.: Glm-4.5v and glm-4.1v-thinking: Towards versatile multimodal reasoning with scalable reinforcement learning. arXiv preprint arXiv:2507.01006  (2025), \url{https://arxiv.org/abs/2507.01006}

\bibitem{gong2019memorizing}
Gong, D., Liu, L., Le, V., Saha, B., Mansour, M.R., Venkatesh, S., Van Den~Hengel, A.: Memorizing normality to detect anomaly: Memory-augmented deep autoencoder for unsupervised anomaly detection. In: ICCV. pp. 1705--1714 (2019)

\bibitem{grauman2022ego4d}
Grauman, K., Westbury, A., Byrne, E., et~al.: Ego4d: Around the world in 3,000 hours of egocentric video. In: CVPR. pp. 18995--19012 (2022)

\bibitem{hasan2016temporal}
Hasan, M., Choi, J., Neumann, J., Roy-Chowdhury, A.K., Davis, L.S.: Learning temporal regularity in video sequences. In: CVPR. pp. 733--742 (2016)

\bibitem{iso22000}
{International Organization for Standardization}: Iso 22000:2018 food safety management systems -- requirements for any organization in the food chain. Tech. rep., ISO (2018)

\bibitem{moonshotai2025kimi}
{Kimi Team}: Kimi k2.5: Visual agentic intelligence. arXiv preprint arXiv:2602.02276  (2026), \url{https://arxiv.org/abs/2602.02276}

\bibitem{li2025anomize}
Li, F., Liu, W., Chen, J., Zhang, R., Wang, Y., Zhong, X., Wang, Z.: Anomize: Better open vocabulary video anomaly detection. In: CVPR. pp. 29203--29212 (2025)

\bibitem{li2024mvbench}
Li, K., Wang, Y., He, Y., Li, Y., Wang, Y., Liu, Y., Wang, Z., Xu, J., Chen, G., Luo, P., Wang, L., Qiao, Y.: Mvbench: A comprehensive multi-modal video understanding benchmark. In: CVPR. pp. 22195--22206 (2024)

\bibitem{lv2023unbiased}
Lv, H., Yue, Z., Sun, Q., Luo, B., Cui, Z., Zhang, H.: Unbiased multiple instance learning for weakly supervised video anomaly detection. In: CVPR. pp. 8022--8031 (2023)

\bibitem{mangalam2024egoschema}
Mangalam, K., Akshulakov, R., Malik, J.: Egoschema: A diagnostic benchmark for very long-form video language understanding. In: NeurIPS (2023)

\bibitem{fda2017haccp}
{National Advisory Committee on Microbiological Criteria for Foods}: Hazard analysis and critical control point principles and application guidelines. Journal of Food Protection  \textbf{61}(9),  1246--1259 (1998). \doi{10.4315/0362-028X-61.9.1246}

\bibitem{park2020memory}
Park, H., Noh, J., Ham, B.: Learning memory-guided normality for anomaly detection. In: CVPR. pp. 14372--14381 (2020)

\bibitem{ramachandra2020streetscene}
Ramachandra, B., Jones, M.J.: Street scene: A new dataset and evaluation protocol for video anomaly detection. In: WACV. pp. 2569--2578 (2020)

\bibitem{sener2022assembly101}
Sener, F., Chatterjee, D., Shelepov, D., He, K., Singhania, D., Wang, R., Yao, A.: Assembly101: A large-scale multi-view video dataset for understanding procedural activities. In: CVPR. pp. 21096--21106 (2022)

\bibitem{stein201450salads}
Stein, S., McKenna, S.J.: Combining embedded accelerometers with computer vision for recognizing food preparation activities. In: ACM Int. Joint Conf. Pervasive Ubiquitous Comput. pp. 729--738 (2013)

\bibitem{sultani2018ucfcrime}
Sultani, W., Chen, C., Shah, M.: Real-world anomaly detection in surveillance videos. In: CVPR. pp. 6479--6488 (2018)

\bibitem{tang2019coin}
Tang, Y., Ding, D., Rao, Y., Zheng, Y., Zhang, D., Zhao, L., Lu, J., Zhou, J.: Coin: A large-scale dataset for comprehensive instructional video analysis. In: CVPR. pp. 1207--1216 (2019)

\bibitem{tian2021rtfm}
Tian, Y., Pang, G., Chen, Y., Singh, R., Verjans, J.W., Carneiro, G.: Weakly-supervised video anomaly detection with robust temporal feature magnitude learning. In: ICCV. pp. 4975--4986 (2021)

\bibitem{wu2020xdviolence}
Wu, P., Liu, J., Shi, Y., Sun, Y., Shao, F., Wu, Z., Yang, Z.: Not only look, but also listen: Learning multimodal violence detection under weak supervision. In: ECCV. pp. 322--339 (2020)

\bibitem{wu2024openvocab}
Wu, P., Zhou, X., Pang, G., Sun, Y., Liu, J., Wang, P., Zhang, Y.: Open-vocabulary video anomaly detection. In: CVPR. pp. 18297--18307 (2024)

\bibitem{wu2024vadclip}
Wu, P., Zhou, X., Pang, G., Zhou, L., Yan, Q., Wang, P., Zhang, Y.: Vadclip: Adapting vision-language models for weakly supervised video anomaly detection. In: AAAI. vol.~38, pp. 6074--6082 (2024)

\bibitem{xiaomi2025mimovl}
{Xiaomi LLM-Core Team}: Mimo-vl technical report. arXiv preprint arXiv:2506.03569  (2025), \url{https://arxiv.org/abs/2506.03569}

\bibitem{yang2024hybridsort}
Yang, M., Han, G., Yan, B., Zhang, W., Qi, J., Lu, H., Wang, D.: Hybrid-sort: Weak cues matter for online multi-object tracking. In: AAAI. vol.~38, pp. 6504--6512 (2024)

\bibitem{yang2022lavt}
Yang, Z., Wang, J., Tang, Y., Chen, K., Zhao, H., Torr, P.H.: Lavt: Language-aware vision transformer for referring image segmentation. In: Proceedings of the IEEE/CVF conference on computer vision and pattern recognition. pp. 18155--18165 (2022)

\bibitem{yang2024textprompt}
Yang, Z., Liu, J., Wu, P.: Text prompt with normality guidance for weakly supervised video anomaly detection. In: CVPR. pp. 18899--18908 (2024)

\bibitem{yu2019activitynetqa}
Yu, Z., Xu, D., Yu, J., Yu, T., Zhao, Z., Zhuang, Y., Tao, D.: Activitynet-qa: A dataset for understanding complex web videos via question answering. In: AAAI. pp. 9127--9134 (2019)

\bibitem{zhipuai2025glm46v}
{Zhipu AI}: Glm-4.6v (2025), \url{https://docs.z.ai/guides/vlm/glm-4.6v}, official documentation page

\bibitem{zhu2025internvl3}
Zhu, J., et~al.: Internvl3: Exploring advanced training and test-time recipes for open-source multimodal models. arXiv preprint arXiv:2504.10479  (2025)

\end{thebibliography}

\end{document}